\documentclass{webofc}
\usepackage[varg]{txfonts}   % Web of Conferences font
\usepackage{xcolor}
\usepackage{graphicx} % remove 'demo' option in real doc.
\usepackage[skip=0.333\baselineskip]{caption}
\usepackage{subcaption}
\usepackage{booktabs}

\usepackage[acronym]{glossaries}
\newacronym{hpo}{HPO}{Hyperparameter Optimization}
\newacronym{hpc}{HPC}{High Performance Computing}
\newacronym{hep}{HEP}{High-Energy Physics}
\newacronym{hp}{HP}{Hyperparameter}
\newacronym{svr}{SVR}{Support Vector Regression}
\newacronym{qsvr}{QSVR}{Quantum Support Vector Regression}
\newacronym{ml}{ML}{Machine Learning}
\newacronym{dl}{DL}{Deep Learning}
\newacronym{mpi}{MPI}{Message Passing Interface}
\newacronym{qa}{QA}{Quantum Annealer}
\newacronym{mlpf}{MLPF}{Machine-Learned Particle Flow}
\newacronym{nn}{NN}{Neural Network}
\newacronym{cnn}{CNN}{Convolutional Neural Network}
\newacronym{lstm}{LSTM}{Long Shot-Term Memory}

\begin{document}

% \title{Optimizing AI-based HEP algorithms \\ using HPC and Quantum Computing}
\title{Model Performance Prediction for Hyperparameter Optimization of Deep Learning Models Using High Performance Computing and Quantum Annealing}


\author{\firstname{Juan Pablo} \lastname{García Amboage}\inst{1,2}\fnsep\thanks{\email{jugarcia@student.ethz.ch}} \and
        \firstname{Eric} \lastname{Wulff}\inst{1}\fnsep\thanks{\email{eric.wulff@cern.ch}} \and
        \firstname{Maria} \lastname{Girone}\inst{1}\and
        \firstname{Tomás} \lastname{F. Pena} \inst{2}
        % etc.
}

\institute{The European Organization for Nuclear Research, CERN 
\and
           CiTIUS, Universidade de Santiago de Compostela, 15782 Santiago de Compostela, Spain
          }

\abstract{%
\gls{hpo} of \gls{dl}-based models tends to be a compute resource intensive process as it usually requires to train the target model with many different hyperparameter configurations. We show that integrating model performance prediction with early stopping methods holds great potential to speed up the \gls{hpo} process of deep learning models. Moreover, we propose a novel algorithm called Swift-Hyperband that can use either classical or quantum \gls{svr} for performance prediction and benefit from distributed \gls{hpc} environments. This algorithm is tested not only for the \gls{mlpf}, model used in \gls{hep}, but also for a wider range of target models from domains such as computer vision and natural language processing. Swift-Hyperband is shown to find comparable (or better) hyperparameters as well as using less computational resources in all test cases.
\glsresetall  % reset acronyms after abstract
}

\maketitle

\section{Introduction}
\label{intro}

Training and \gls{hpo} of \gls{dl} models is often compute resource intensive and calls for the use of large-scale \gls{hpc} resources as well as scalable and resource efficient \gls{hp} search and evaluation algorithms \cite{MlpfHPO}. Current state-of-the-art \gls{hpo} algorithms such as Hyperband \cite{hyperband}, ASHA \cite{asha}, and BOHB \cite{bohb}, rely on a method of early termination. Badly performing trials are automatically terminated to allocate compute resources to more promising ones.
Such methods have been successfully applied to optimize \gls{mlpf}, a particle flow reconstruction \gls{nn} \cite{mlpf1}. Using this technique led to o a reduction of $\sim$44\% in the validation loss of MLPF \cite{MlpfHPO}.\\

In this context, performance prediction emerges as a potential approach to accelerate the \gls{hpo} process. This involves using meta-models, referred to as a performance predictors, that can estimate the performance of a given configuration at a particular epoch by leveraging information from its partial learning curve. By employing performance prediction, it is possible to prioritize the training of the most promising configurations based on their predicted performance, while avoiding the need to fully train configurations with poorer predicted performance. Consequently, this approach holds great potential for reducing the time and computational resources required for the \gls{hpo} process.\\

This work explores novel techniques based on performance prediction to accelerate the \gls{hpo} process of \gls{mlpf} and other \gls{nn} architectures that leverage the use of \gls{hpc} resources for training the target model and quantum computing for training the performance predictors. Moreover, a new \gls{hpo} algorithm is proposed, Swift-Hyperband, that integrates Hyperband with the use of model performance predictors.

\section{Related work}
Baker et al. \cite{perfpred} demonstrated that \gls{svr} models can effectively serve as performance predictors for various \gls{nn} architectures. In addition to showing a good predictive capability for performance prediction tasks, \gls{svr} models offer the advantage of having negligible training and inference times, even when using a consumer-grade laptop CPU. Hence, using \glspl{svr} prevents the training of performance predictors from becoming a bottleneck for the potential resource savings expected from this technique.\\

In the European Center of Excellence in Exascale Computing "Research on AI- and Simulation-based Engineering at Exascale" (CoE RAISE) the capability of \glspl{svr} to predict the loss of \gls{mlpf} after 100 training epochs in the Delphes dataset \cite{data_delphes} was successfully shown \cite{acat}, achieving $R^2$ values of around 0.9 when using 25\% of the target learning curve as input. Here, $R^2$ is the so-called coefficient of determination, defined as

\begin{equation}
R^2 = 1-\frac{\sum_{i} (y_i - f_i)^2}{\sum_{i} (y_i - \Bar{y})^2} = 1-\frac{\sum_{i} e_i^2}{\sum_{i} (y_i - \Bar{y})^2}
\end{equation}

where $y_i$ is the ground truth for data point $i$, $f_i$ is prediction $i$, $\Bar{y}$ is the mean of all $y_i$ and $e_i$ is the error of prediction $i$. Furthermore, the Quantum Annealer at the Jülich Supercomputer Centre, was used to train \gls{qsvr} \cite{pasetto} models on \gls{mlpf} learning curves. While no significant performance benefit was expected from the use of quantum resources, and the idea of employing quantum computers for the task of performance prediction was primarily a proof of concept of integrating this technology into the \gls{hpo} workflow, the \glspl{qsvr} achieved comparable performance to that obtained with classical \glspl{svr} \cite{acat}. Note that the number of training samples in this case had to be reduced to 20 due to limitations in problem size arising from the current state of quantum technologies. \\

There are several strategies that can be considered for integrating performance predictors with the \gls{hpo} process \cite{perfpred}. A straightforward approach is to generate a certain amount of random \gls{hp} configurations, fully train $M$ configurations and partially train $N>M$ configurations. The final loss of the fully trained configurations and part of their learning curves are used to train a performance predictor. Then, this performance predictor is used to predict the final loss of the partially trained configurations and only those whose predicted loss is below a certain threshold are selected to complete training. This approach was tested using different types of performance predictors, including \glspl{qsvr}, in \cite{aachhybrid}.\\

Another, more sophisticated example of the use of performance predictors for HPO is the algorithm Fast-Hyperband \cite{perfpred}. This algorithm is a modified version of the well-known Hyperband algorithm that uses performance predictors which are trained on the fly during the execution of the algorithm to save training epochs of the target model with respect to Hyperband. More precisely, Fast-Hyperband adds an extra decision point \footnote{We understand a decision point as a certain epoch in which an early stopping HPO algorithm discards some configurations to keep training only the most promising ones according to a certain criterion.} based on performance prediction for every epoch in each Hyperband round. Decisions in these new intermediate points use a probability threshold, computed from an estimate of the standard deviation of every predictor used. The proposed method of computing this estimate in Fast-Hyperband is leave one out cross validation. The main drawback of this approach is that it requires to train many performance predictors, which makes it impractical to use \glspl{qsvr}. This is partly due to runtime limitations on the quantum machine and partly due to the time needed to formulate the regression problem in a suitable way for the quantum annealer. In addition, the time spent to connect to the quantum machine needs to be taken into account. Furthermore, Fast-Hyperband, as it is defined in \cite{perfpred}, is a sequential algorithm which makes it unable to benefit from running in a distributed manner on multiple nodes in an HPC environment. 

\section{Swift-Hyperband}
\begin{figure}[h!]
    \centering
    \includegraphics[width=10cm]{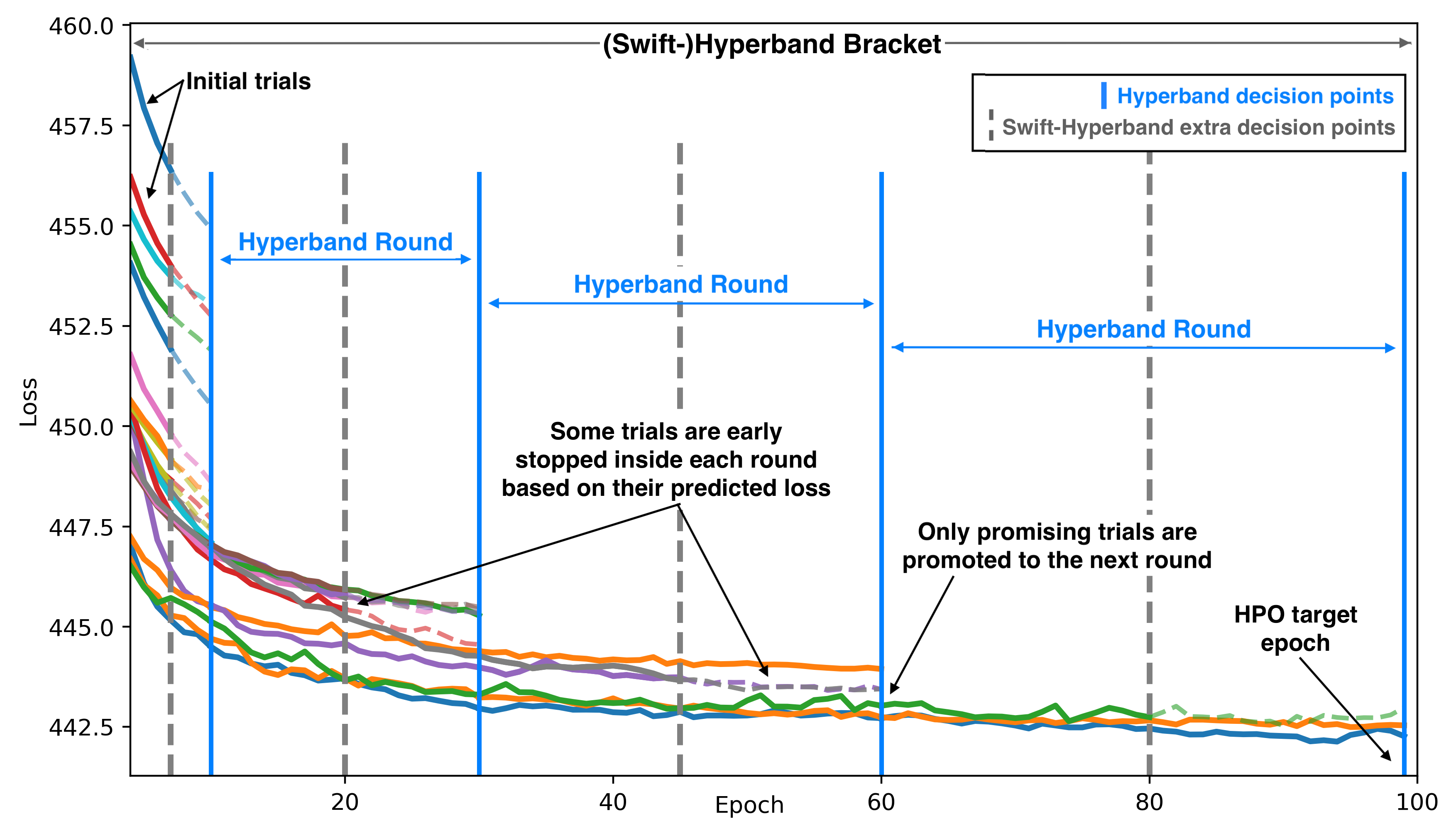}
    \caption{Graphic representation of a Swift-Hyperband bracket. This is an illustrative example, not the result of an actual execution the algorithm.}
    \label{fig: s-hb}
\end{figure}

To overcome the limitations of Fast-Hyperband mentioned in the previous section we propose a different way to integrate  performance predictors with Hyperband by a new algorithm called Swift-Hyperband. Similarly to Fast-Hyperband, Swift-Hyperband adds extra decision points to the classical Hyperband algorithm that are based on performance prediction. However, Swift-Hyperband adds only one extra decision point per Hyperband round. When a round starts, Swift-Hyperband trains a certain number of trials until the end of the round and then uses their loss at the end of the round to define a threshold. The other trials in the round are partially trained to the extra decision point, where the performance predictors are used to predict their losses at the end of the round. These predictions are compared to the previously defined threshold to decide if each trial should be terminated or not.  \\ 

Figure \ref{fig: s-hb} illustrates what happens inside each bracket of Swift-Hyperband. The vertical dashed lines represent the new decision points in which some configurations or trials are discarded based on their predicted performance. The trials that did not complete their round are represented using dashed learning curves. That is, the dashed learning curves represent training epochs of the target model that Swift-Hyperband saved with respect to the classical Hyperband algorithm. The trials that complete their training until the end of the round are the ones that are always represented by continuous lines. The decision of which of these trials are promoted to the next round is made as it is made in the classical Hyperband algorithm.

\begin{figure}[h]
    \centering
    \includegraphics[width=10cm]{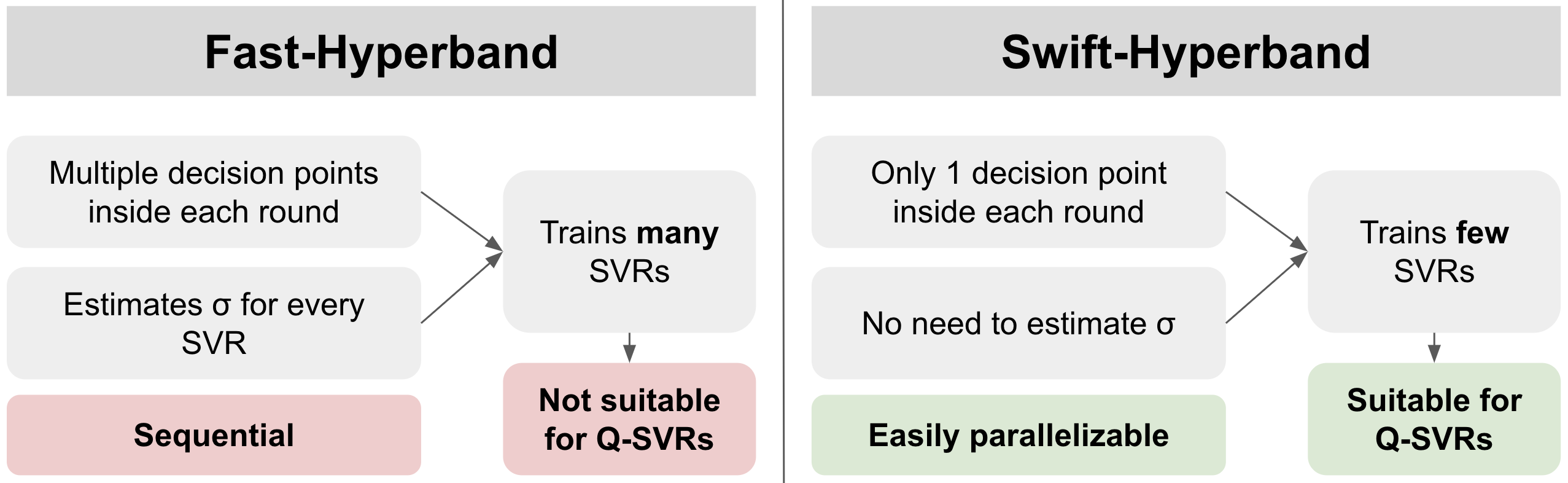}
    \caption{Key differences between Fast-Hyperband and Swift-Hyperband.}
    \label{fig: compa}
\end{figure}

Swift-Hyperband not only needs to train less performance predictors than Fast-Hyperband but can also be easily parallelized, as all the initial full and partial trainings inside a round can be executed in parallel. For these two reasons, our algorithm can potentially use \glspl{qsvr} and benefit  from HPC environments. A schematic comparison between Fast-Hyperband and Swift-Hyperband is shown in Figure \ref{fig: compa}.

\section{Results}
To compare Hyperband, Fast-hyperband, Swift-Hyperband and Quantum-Swift-Hyperband (Swift-Hyperband using \glspl{qsvr}) for different \gls{nn} architectures we simulate 10 runs of each algorithm using the datasets of learning curves derived from the following model-dataset combinations:

\begin{itemize}
    \item \gls{mlpf} \cite{acat} trained on the Delphes dataset \cite{data_delphes}
    \item An image recognition \gls{cnn} modified from \cite{asha} trained on the Cifar10 dataset
    \item An image recognition \gls{cnn} trained on the SVHN dataset used in \cite{perfpred}
    \item A natural language processing \gls{lstm} \gls{nn} trained in the PTB dataset 
\end{itemize}

\begin{figure}[hbt!]
\begin{subfigure}{.475\linewidth}
  \includegraphics[width=\linewidth]{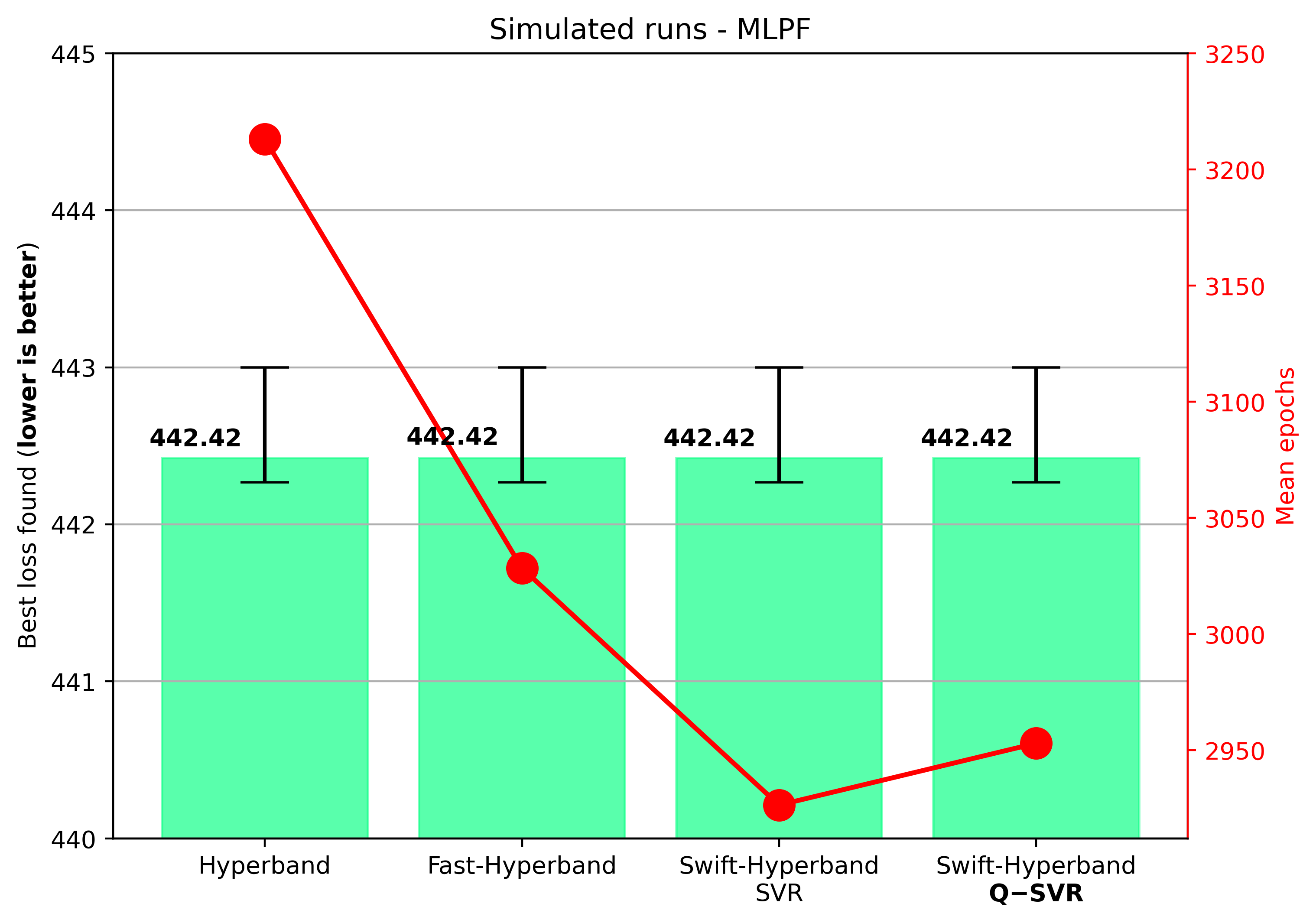}
  \caption{MLPF trained on Delphes}
  \label{MLEDdet}
\end{subfigure}\hfill % <-- "\hfill"
\begin{subfigure}{0.475\linewidth}
  \includegraphics[width=\linewidth]{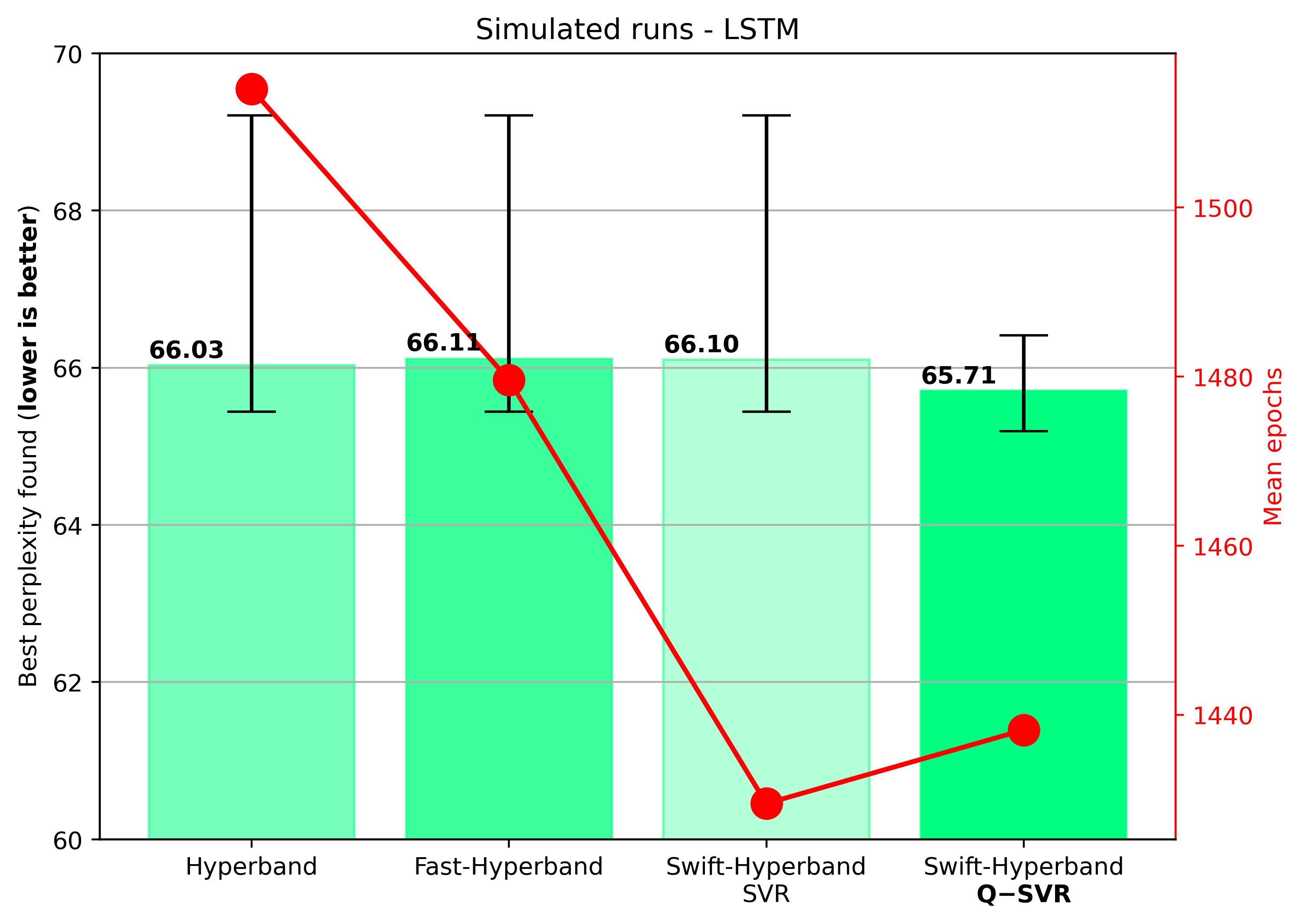}
  \caption{\gls{lstm} trained on PTB}
  \label{energydetPSK}
\end{subfigure}

%\medskip % create some *vertical* separation between the graphs
\begin{subfigure}{.475\linewidth}
  \includegraphics[width=\linewidth]{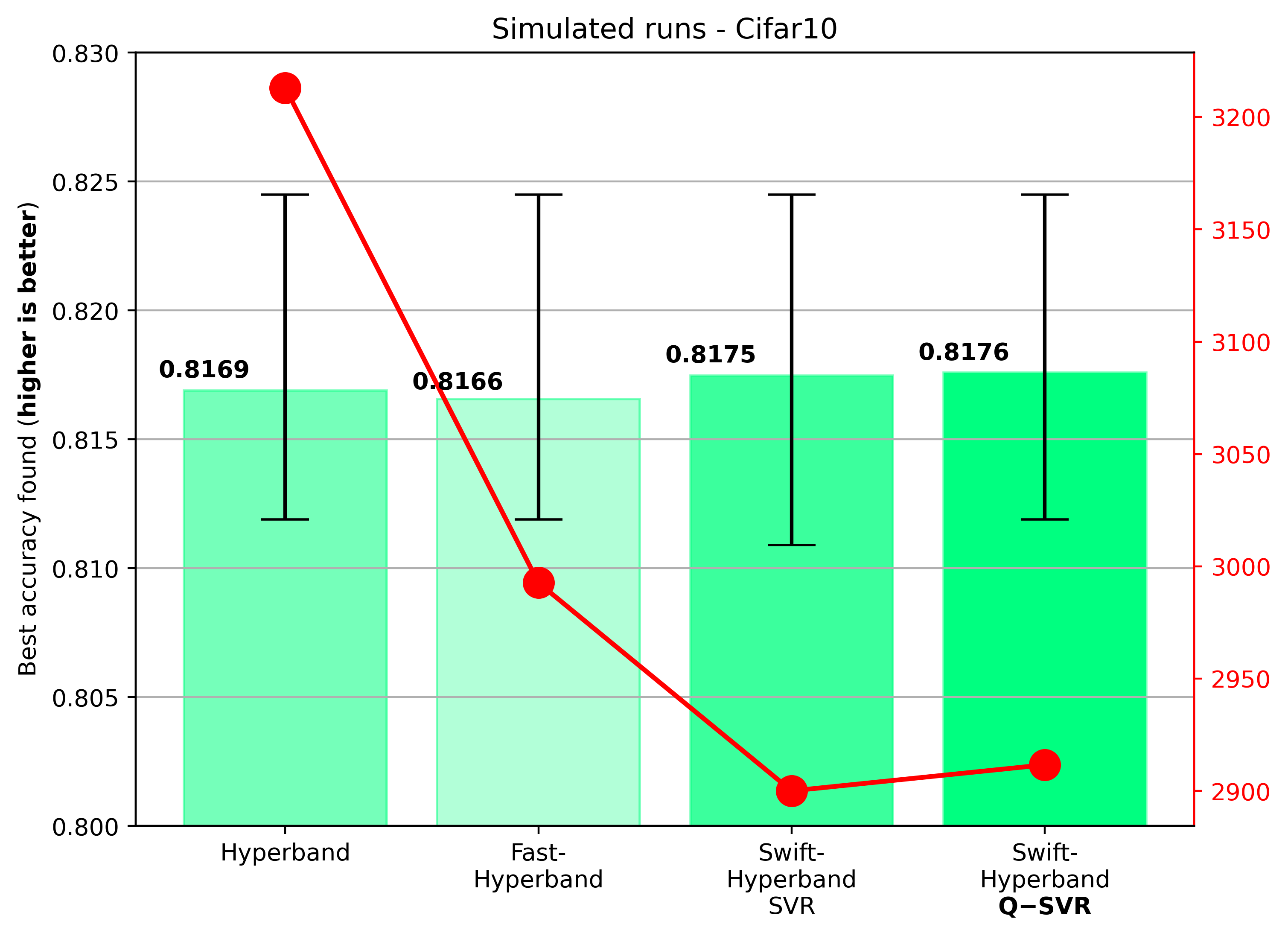}
  \caption{\gls{cnn} trained on Cifar10}
  \label{velcomp}
\end{subfigure}\hfill % <-- "\hfill"
\begin{subfigure}{0.475\linewidth}
  \includegraphics[width=\linewidth]{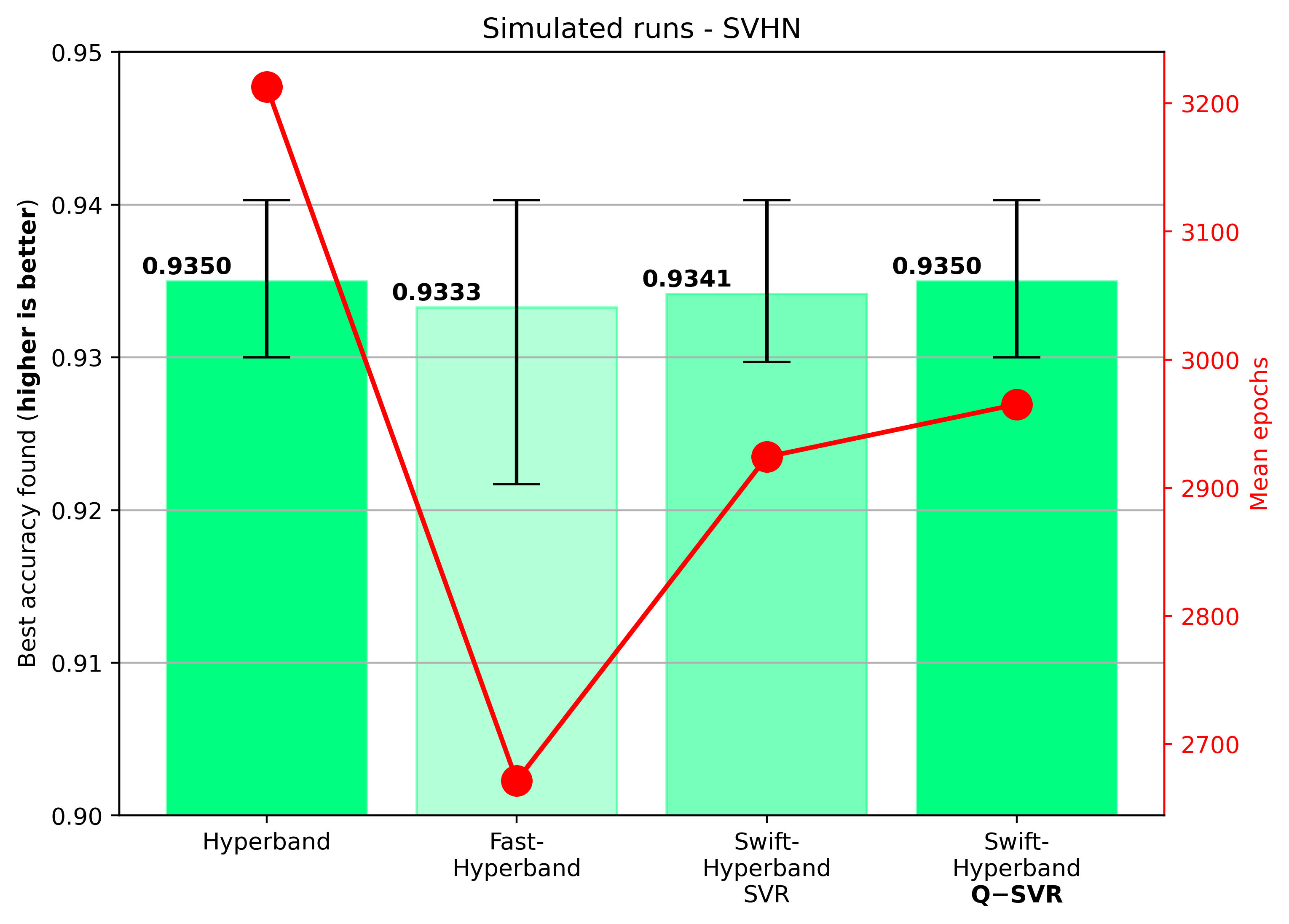}
  \caption{\gls{cnn} trained on SVHN}
  \label{estcomp}
\end{subfigure}

\textbf{}\caption{Average, best, and worst performance of the best configuration found as well as the total number of epochs consumed by the different HPO algorithms for different \glspl{nn} and datasets.}
\label{fig:sim}
\end{figure}

% The first is made up of training cycles of \gls{mlpf} \cite{acat}, the second is based on an image recognition \gls{cnn} modified from \cite{asha} trained on the Cifar10 dataset and an image recognition \gls{cnn} trained on the SVHN dataset used in \cite{perfpred} and lastly the natural language processing \gls{lstm} \gls{nn} trained in the PTB dataset that is used in \cite{perfpred}.
The result of these simulated runs can be seen in Figure \ref{fig:sim} and a summary of the learning curve datasets used for the simulation is available in Table \ref{lcds}.\\

\begin{table}[]
\centering
\begin{tabular}{@{}cccc@{}}
\toprule
\tiny
\textbf{Neural Network} & \textbf{Evluation metric} & \textbf{\begin{tabular}[c]{@{}c@{}}HPs search space \\ dimension\end{tabular}} & \textbf{\begin{tabular}[c]{@{}c@{}}Target epoch\\ for HPO\end{tabular}} \\ \midrule
MLPF for Delphes        & Loss                      & 7                                                                              & 100                                                                     \\
LSTM for PTB            & Perplexity                & 2                                                                              & 60                                                                      \\
\gls{cnn} for Cifar10         & Accuracy                  & 5                                                                              & 100                                                                     \\
\gls{cnn} for SVHN            & Accuracy                  & 9                                                                              & 100                                                                     \\ \bottomrule
\end{tabular}
\caption{Summary of learning curve datasets.}
\label{lcds}
\end{table}

\begin{figure}[h!]
    \centering
    \includegraphics[width=.55\linewidth]{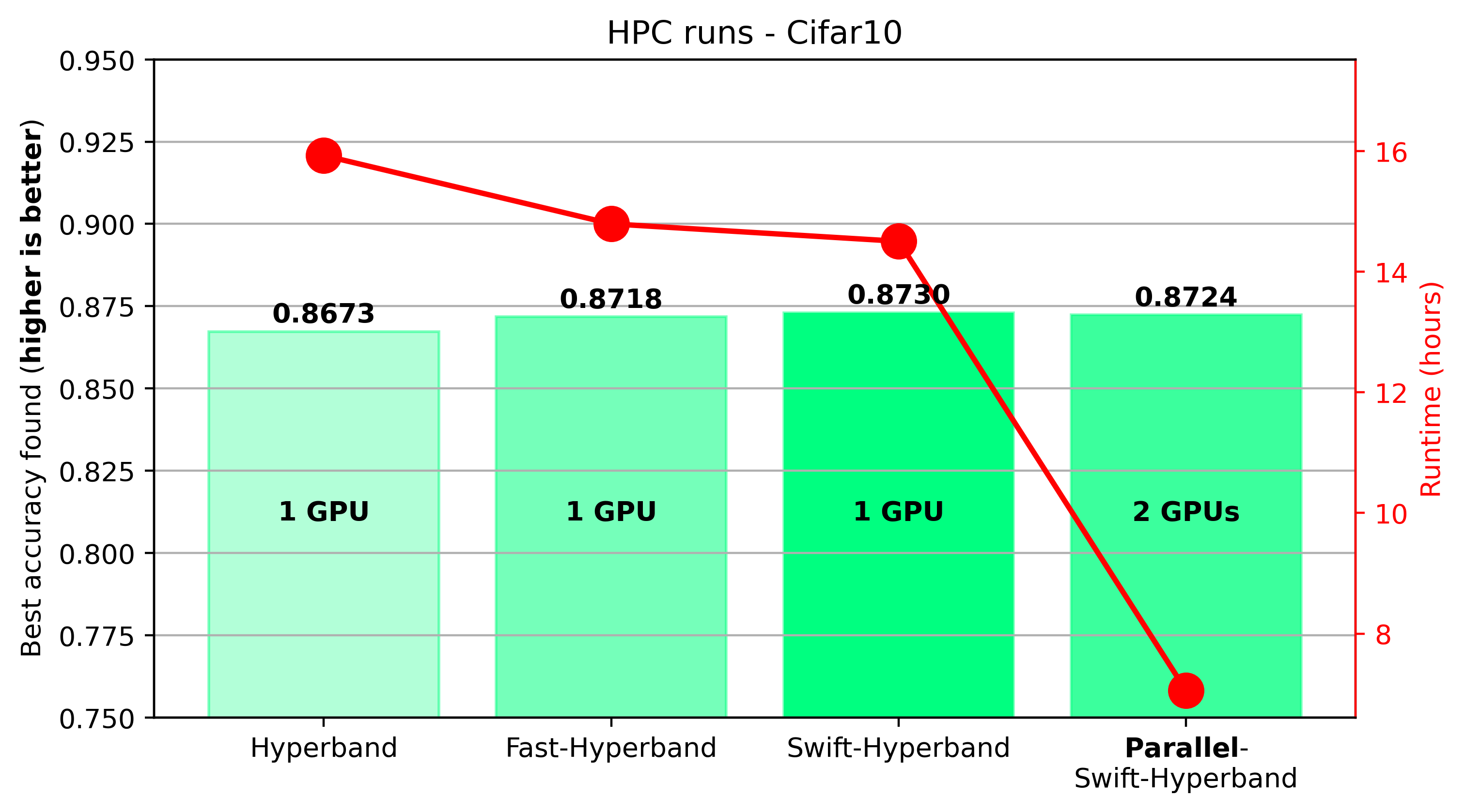}
    \caption{Performance of the best configuration found and total runtime needed for different HPO algorithms.}
    \label{fig:cifar_real}
\end{figure}

Beyond the simulated runs, we test the speedup provided by the parallelization of Swift-Hyperband along with the achieved accuracies by running Hyperband, Fast-Hyperband, Swift-Hyperband, and a parallel version of Swift-Hyperband that uses MPI to coordinate one CPU node and two GPU worker nodes. For these runs, the HPO target was a simple 6-layer \gls{cnn} (different to the \gls{cnn} used in the simulated runs) trained on Cifar10 using a 3-dimensional search space consisting of learning rate, weight decay, and dropout. This network was chosen because it was relatively fast to train. The results can be seen in Figure \ref{fig:cifar_real}.

The results in Figure \ref{fig:sim} and Figure \ref{fig:cifar_real} show that both Swift-Hyperband and its version using \glspl{qsvr} achieve accuracies comparable to classical Hyperband while needing considerably fewer epochs in all cases. In comparison to Fast-Hyperband, Swift-Hyperband (SVR and Q-SVR) is faster in all cases except on the SVHN problem. When it comes to the non-simulated runs we observe that all algorithms achieve accuracies around 87$\%$, with both Swift-Hyperband and Parallel-Swift-Hyperband slightly beating Fast-Hyperband. Note that in several cases the version of Swift-Hyperband that uses \glspl{qsvr} finds better performing configurations than the original Hyperband algorithm, something that would not happen if the \glspl{qsvr} made perfect predictions. This may indicate that using performance predictors, aside from saving compute resources, has some type of regularizing effect that prevents some of the errors that Hyperband makes when terminating configurations at the end of each round.

\section{Conclusions}
We proposed a new promising parallelizable \gls{hpo} algorithm integrating Hyperband and performance predictors that can be used in combination with \glspl{svr} or \glspl{qsvr}. This leaves the door open for the use of Swift-Hyperband in later \gls{hpo} cycles of \gls{mlpf}. Furthermore, it was shown that, despite the current limitations of quantum computers, it is possible to execute hybrid Quantum/HPC workflows for HPO, achieving comparable performance to fully classical workflows. We consider that there is a need for further studies on the speedup achieved by the parallelization of Swift-Hyperband when using a greater number of nodes as Hyperband is known to suffer from straggler issues \cite{asha}. Hence, developing a version of the HPO algorithm ASHA that integrates performance predictors, potentially to be named Swift-ASHA, is one of the identified lines to continue this work. In addition to this, conducting more empirical tests of Swift-Hyperband on a wider range of target models would provide valuable insights on the behavior of the algorithm. Finally, conducting theoretical studies on \glspl{svr} and \glspl{qsvr} could provide a deeper understanding of why Swift-Hyperband occasionally outperforms the original Hyperband algorithm and shed light on the differences observed when employing quantum or classical SVRs.

\section*{Acknowledgements}
Eric Wulff and Juan Pablo García Amboage was supported by CoE RAISE. The CoE RAISE project has received funding from the European Union’s Horizon 2020 – Research and Innovation Framework Programme H2020-INFRAEDI-2019-1 under grant agreement no. 951733.

The authors gratefully acknowledge the computing time granted through JARA on the supercomputer JURECA \cite{jureca} at Forschungszentrum Jülich. The authors gratefully acknowledge the Jülich Supercomputing Centre for funding this project by providing computing time through the Jülich UNified Infrastructure for Quantum computing (JUNIQ) on the D-Wave Advantage™ System JUPSI.

\bibliography{bibliography}

\end{document}